\documentclass[conference]{IEEEtran}

\usepackage[letterpaper, left=0.680in, right=0.680in, bottom=1.049in, top=0.71in]{geometry}

\usepackage{ifpdf}
\ifCLASSINFOpdf
\usepackage[pdftex]{graphicx}
\graphicspath{{./Figures/}}
\DeclareGraphicsExtensions{.pdf,.jpeg,.png}
\else
\usepackage[dvips]{graphicx}
\DeclareGraphicsExtensions{.pdf}
\usepackage[caption=false, font=footnotesize]{subfig}
\fi

\usepackage{tabularx, booktabs}
\usepackage{amssymb}
\usepackage{amsthm} 
\usepackage[ruled,vlined,linesnumbered]{algorithm2e}
\usepackage{aligned-overset} 
\usepackage{array}
\usepackage{balance}
\usepackage{cite}
\usepackage{color}
\usepackage{epstopdf}
\usepackage{enumitem} 
\usepackage{mathtools} 
\usepackage{makecell} 
\usepackage{multicol} 
\usepackage{multirow} 
\usepackage[caption=false,font=small]{subfig}
\usepackage{textcomp}
\usepackage{stfloats}
\usepackage{url}
\usepackage{verbatim}
\usepackage{graphicx}
\usepackage{url}
\usepackage{float}
\usepackage{xcolor}

\definecolor{cyan}{RGB}{0, 102, 204} %
\definecolor{red}{RGB}{204, 0, 0} %
\definecolor{green}{RGB}{5,107,68} %
\definecolor{purple}{RGB}{102, 51, 153} %
\definecolor{blue}{rgb}{0.06, 0.2, 0.65}
\usepackage{hyperref}
\hypersetup{colorlinks=true,linkcolor=red,urlcolor=blue, citecolor=blue}

\newtheorem{remark}{Remark}

\begin{document}

\title{Feature Fusion and Knowledge-Distilled Multi-Modal Multi-Target Detection}
\author{\IEEEauthorblockN{
            Ngoc~Tuyen~Do\IEEEauthorrefmark{1} and 
            Tri~Nhu~Do\IEEEauthorrefmark{2}%
			}
		\IEEEauthorblockA{
		\IEEEauthorrefmark{1}School of Information and Communications, Hanoi University of Science and Technology\\
        \IEEEauthorrefmark{2}Telecom Neural Detection Lab, Polytechnique Montr\'{e}al, Montreal, QC, Canada
        }
		\IEEEauthorblockA{ Emails: 
			tuyen.dn242305m@sis.hust.edu.vn,
			tri-nhu.do@polymtl.ca
			}
	}
\maketitle
\begin{abstract}
In the surveillance and defense domain, multi-target detection and classification (MTD) is considered essential yet challenging due to heterogeneous inputs from diverse data sources and the computational complexity of algorithms designed for resource-constrained embedded devices, particularly for AI-based solutions. To address these challenges, we propose a feature fusion and knowledge-distilled framework for multi-modal MTD that leverages data fusion to enhance accuracy and employs knowledge distillation for improved domain adaptation. Specifically, our approach utilizes both RGB and thermal image inputs within a novel fusion-based multi-modal model, coupled with a distillation training pipeline. We formulate the problem as a posterior probability optimization task, which is solved through a multi-stage training pipeline supported by a composite loss function. This loss function effectively transfers knowledge from a teacher model to a student model. Experimental results demonstrate that our student model achieves approximately 95\% of the teacher model’s mean Average Precision while reducing inference time by approximately 50\%, underscoring its suitability for practical MTD deployment scenarios.
\end{abstract}

\begin{IEEEkeywords}
Mutli-target detection, knowledge distillation, feature fusion, optimization, AI/ML, FLIR, thermal data, RGB
\end{IEEEkeywords}

\section{Introduction}

Multi-target detection (MTD), which aims to identify and classify multiple targets simultaneously, is a pivotal task in applications such as surveillance, autonomous systems, and radar-based tracking. Traditional MTD approaches predominantly relied on hand-crafted feature engineering coupled with statistical models. For instance, the Histogram of Oriented Gradients (HOG) features combined with Support Vector Machine (SVM) classifiers \cite{HOG_SVM} were widely adopted for object detection and classification tasks prior to the advent of deep learning, with numerous algorithmic variants achieving notable success.
Deep learning has significantly enhanced MTD capabilities by enabling the automatic extraction of rich, hierarchical, and non-linear features directly from raw data. In particular, Convolutional Neural Networks (CNNs) serve as the foundational architecture for state-of-the-art (SOTA) object detection models, such as the EfficientDet family \cite{efficientdet} and various iterations of MobileNet \cite{mobilenetv3}.

Several challenges in MTD must be overcome, including aligned data fusion from heterogeneous inputs (e.g., RGB, thermal), large model sizes that deter response of edge embedding devices, and limited generalization to difficult environments. On the one hand, deep learning Knowledge Distillation (KD) approach \cite{kd_od} is a training technique where a large, accurate \textit{teacher model} conveys pre-trained knowledge to a compact \textit{student model} without significant performance degradation. Unlike single-output tasks, MTD produces multiple bounding boxes accompanied by labels and scores, which makes it challenging to define the posterior problem formulation and determine the knowledge to be transferred (e.g., classification logits, localization features). To address the complexity of the training procedure, the loss function of the proposed method requires careful architectural and algorithmic design to produce effective distillation.
On the other hand, \textit{multi-modal} sensor fusion has been explored \cite{bijelic2020fusion} to enhance MTD by integrating complementary information from diverse input sensors, addressing adverse weather conditions, such as RGB and thermal. DeepInversion for Object Detection (DIODE) \cite{chawla2021kd} developed a data-free KD model with significant improvements, including data augmentations and an automated bounding box and category sampling scheme. CrossKD \cite{wang2024crosskd} is a novel KD method that significantly enhances the AP of GFL ResNet-50, where the intermediate feature maps of the student model are conveyed to the teacher to receive contradictory supervision signals.

\textit{In this paper}, we investigate the MTD problem in the context of autonomous driving surveillance, using realistic datasets. We aim to address several technical challenges mentioned above. To this end, we propose a \textit{Feature Fusion and Knowledge-Distilled} (FFKD) method, which refers to a framework that integrates multi-modal data or features (fusion) and employs KD to transfer knowledge from a foundation (teacher) model to a simpler (student) model. Our contributions are twofold: (i) we provide a publicly accessible repository of implementation code,\footnote{The code repository can be accessed at: \url{https://github.com/TND-Lab/Feature-Fusion-Knowledge-Distilled-Multi-Modal-Multi-Target-Detection}} and (ii) our technical contributions are detailed as follows:
\begin{itemize}
    \item We formulate a posterior distribution-based optimization problem to rigorously characterize the MM-MTD task.
    \item We propose a multi-modal model combining a fusion method for diverse inputs and a distillation composite loss function to solve the formulated problem.
    \item The experimental results demonstrate that our approach effectively produces a lightweight model for the considered MTD scenario with heterogeneous inputs in challenging conditional environments.
    \item Specifically, our student model achieves approximately 95\% of the teacher model’s performance in terms of mAP index while offering approximately 50\% faster inference time.
\end{itemize}

\section{System Description}

In the context of MTD, we consider a sophisticated sensing system designed to operate within a dynamic urban environment, tasked with the simultaneous detection and classification of multiple mobile targets, which is described as follows. 

\subsection{Multi-modal input-based MTD}

The system integrates a hybrid sensing device comprising a thermal camera and an RGB camera, both co-located at coordinates \( (x_S, y_S, z_S) \).
The thermal camera, characterized by a resolution of \( W \times H \) pixels, captures heat emissions to produce a thermal image \( I^{\text{thm}} \in \mathbb{R}^{W \times H} \), where each pixel’s intensity corresponds to the thermal radiation of targets within the scene. 
Similarly, the RGB camera, with an identical or comparable resolution of \( W \times H \) pixels, captures visible light to produce an RGB image \( I^{\text{rgb}} \in \mathbb{R}^{W \times H \times 3} \), where each pixel encodes color information in three channels (red, green, blue). 

The system processes a multi-modal input \( (I^{\text{thm}}, I^{\text{rgb}}) \), where features from the thermal and RGB images are combined to enhance detection and classification performance. The scene contains \( N \) targets (e.g., pedestrians, vehicles, traffic signs), each characterized by a 2D bounding box \( [x_i, y_i, w_i, h_i] \) in the image plane, where \( [x_i, y_i] \) denotes the top-left corner and \( w_i, h_i \) represent the width and height, respectively. Each target is assigned a categorical label \( c_i \in \{1, 2, \ldots, C\} \), where \( C \) is the number of classes (e.g., \( C = 3 \) for categories such as person, car, and bike). The ground-truth annotations for a given frame are represented as
\begin{align}
A = \{[x_i, y_i, w_i, h_i, c_i]\}_{i=1}^N
\end{align}
where \( N \) varies based on the number of targets present.

The system employs a deep learning model to process the multi-modal input and generate predicted annotations, formalized as
\begin{align}
\hat{A} = f(I^{\text{thm}}, I^{\text{rgb}}; \theta)
\end{align}
where \( f \) is a parameterized model with weights \( \theta \), and the predicted annotations are
\begin{align}
    \label{eq_output_full}
\hat{A} = \{[\hat{x}_i, \hat{y}_i, \hat{w}_i, \hat{h}_i, \hat{p}_i(c)]\}_{i=1}^{\hat{N}}
\end{align}
Here, \( [\hat{x}_i, \hat{y}_i, \hat{w}_i, \hat{h}_i] \) encapsulates the predicted bounding box for the \( i \)-th target, \( \hat{p}_i(c) \) represents the class probabilities across \( C \) classes, and \( \hat{N} \) is the number of predicted targets. The model leverages a multi-modal feature engineering framework, as outlined in the contributions, to fuse thermal and RGB features, enhancing the accuracy of a lightweight model through a distillation training approach inspired by a foundation (teacher) model.

\subsection{Probabilistic Characterization of MM-MTD}

The multi-modal Multi-Target (MM)-MTD problem entails the simultaneous localization and categorization of multiple targets within a dynamic scene, observed by a multi-modal sensing system comprising a thermal camera and an RGB camera, as aforementioned.

\begin{remark}
   The considered MM-MTD is formulated as finding the most accurate approximation of the ground-truth posterior probability of the hypothesis (the set of annotations \( A \)) given the evidence (the thermal and RGB images, \( I^{\text{thm}} \) and \( I^{\text{rgb}} \)), i.e., $P(\hat{A} \mid I^{\text{thm}}, I^{\text{rgb}}) \approx P(A \mid I^{\text{thm}}, I^{\text{rgb}})$.
\end{remark}

The probabilistic formulation seeks the \textit{posterior probability} \( P(A \mid I^{\text{thm}}, I^{\text{rgb}}) \), which quantifies the likelihood of the annotations \( A \) given the evidence provided by the thermal and RGB images. Using Bayes’ theorem, the objective is to infer the set of annotations \( A = \{[b_i, c_i]\}_{i=1}^N \), modeling the joint posterior distribution
\begin{align}
P(A \mid I^{\text{thm}}, I^{\text{rgb}}) = \frac{P(I^{\text{thm}}, I^{\text{rgb}} \mid A) P(A)}{P(I^{\text{thm}}, I^{\text{rgb}})},
\end{align}
where \( P(I^{\text{thm}}, I^{\text{rgb}} \mid A) \) is the \textit{likelihood}, modeling the probability of observing the images given the annotations;
\( P(A) \) is the \textit{prior}, capturing prior knowledge about the annotations (e.g., number of targets, bounding box distributions, class frequencies); and \( P(I^{\text{thm}}, I^{\text{rgb}}) \) is the \textit{evidence}, a normalizing constant ensuring the posterior is a valid probability distribution.

The objective of the detection problem is to find the most likely set of annotations \( A \), typically by maximizing the posterior probability, i.e., Maximum A Posteriori (MAP) estimation, or by characterizing the full posterior distribution for probabilistic inference.

\subsubsection{Joint Likelihood Probability of Multi-Modal Inputs}

\begin{figure}[!ht]
\centering
    \subfloat[RGB Images]{
        \includegraphics[width=0.45\linewidth]{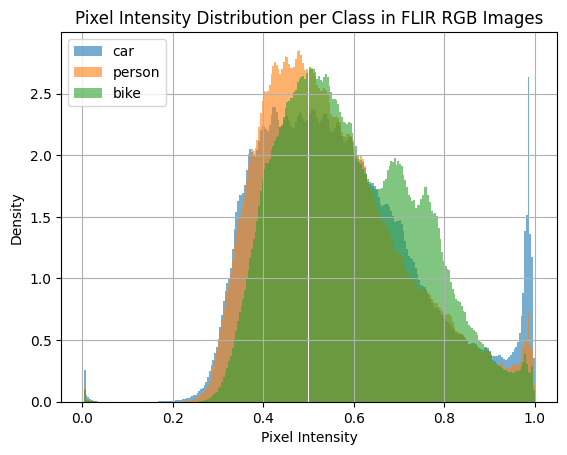}
        \label{pixel_intensity_dist_rgb}
    }
    \hfill
    \subfloat[Thermal Images]{
        \includegraphics[width=0.45\linewidth]{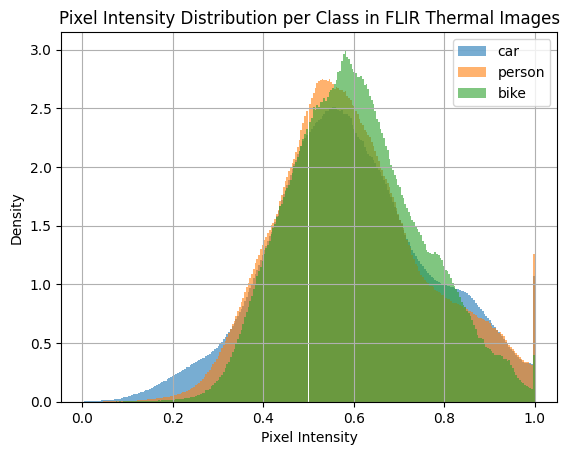}
        \label{pixel_intensity_dist_thermal}
    }
\caption{Pixel intensity distribution per class in (a) RGB images and (b) thermal images in the utilized dataset \cite{dataset_flir_aligned}.}
\label{fig_pixel_intensity}
\end{figure}

Assuming conditional independence between thermal and RGB observations given the annotations, the likelihood factorizes as
\begin{align}
P(I^{\text{thm}}, I^{\text{rgb}} \mid A) = P(I^{\text{thm}} \mid A) P(I^{\text{rgb}} \mid A).
\end{align}
The likelihood for each modality models pixel values as conditionally independent given the target annotations as
\begin{align}
P(I^{\text{img}} \mid A) = \prod_{k \in \Omega} P(I^{\text{img}}(k) \mid A),
\end{align}
where \( {\text{img}} \in \{\text{thm}, \text{rgb}\} \) denotes the modality, \( \Omega \) is the set of pixels, \( k \) indexes a pixel at \( (u_k, v_k) \), \( I^{\text{thm}}(k) \) represents the thermal intensity, and \( I^{\text{rgb}}(k) \in \mathbb{R}^3 \) represents the RGB color vector at pixel \( k \). 
It is noted that \( P(I^{\text{thm}} \mid A) \) and \( P(I^{\text{rgb}} \mid A) \) can be numerically determined, e.g., using the KDE method, as illustrated in Fig.~\ref{fig_pixel_intensity} for the considered FLIR dataset \cite{dataset_flir_aligned}.

\subsubsection{Prior Distribution of the Desired MTD Output}

\begin{figure}[!ht]
\centering
    \includegraphics[width=0.6\linewidth]{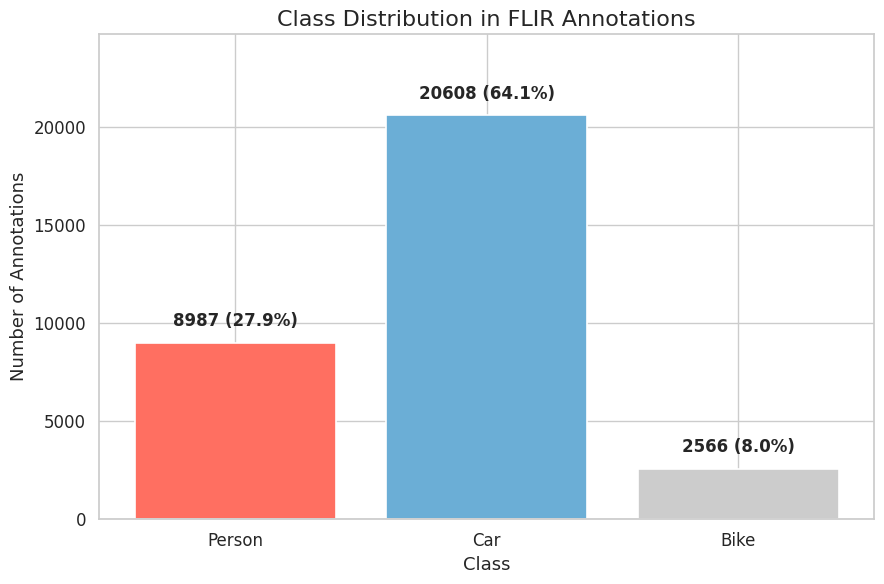}
    \caption{Class distribution in the utilized dataset \cite{dataset_flir_aligned}.}
    \label{fig_class_distribution}
\end{figure}

The prior distribution \( P(A) \) models the number of targets, their bounding boxes, and class labels as
\begin{align}
P(A) = P(N) \prod_{i=1}^N P(b_i) P(c_i).
\end{align}
where $N$ is the number of targets \( N \); bounding boxes are uniformly distributed over the image plane as
$
P(b_i) = \frac{1}{W H W_{\text{max}} H_{\text{max}}},
$
where \( W_{\text{max}}, H_{\text{max}} \) are maximum dimensions; class labels follow a categorical distribution as
$
P(c_i) = \pi_{c_i},
$
where \( \pi_{c_i} \) reflects the expected class frequency, as illustrated in Fig.~\ref{fig_class_distribution}.


\subsection{Optimization Problem Formulation for FFKD in MM-MTD}
\label{distillation_realtime_mtd}

During distillation training, the student model learns to imitate the teacher's predictions by utilizing the teacher’s soft labels, which encode richer inter-class relationships than one-hot ground-truth labels \cite{kd_od}. Additionally, to capture contextual knowledge, the student model replicates the intermediate feature maps of the teacher model. The optimization training objective combines two loss components a KD loss \cite{kd_od}, which measures the difference between the student and teacher outputs, and a feature distillation (FD) loss \cite{feature_distillation}, which aligns the student’s internal representations with those of the teacher.

With a dataset \( \mathcal{D} = \{(I^{thm,(k)}, I^{rgb,(k)}, A^{(k)})\}_{k=1}^K \), where \( A^{(k)} = \{[b_i^{(k)}, c_i^{(k)}]\}_{i=1}^{N^{(k)}} \) are the ground-truth annotations for the \( k \)-th sample, the MM-MTD problem involves inferring a set of annotations \( \hat{A}^{(k)} = \{[\hat{b}_i^{(k)}, \hat{c}_i^{(k)}]\}_{i=1}^{N^{(k)}} \), where \( \hat{b}_i^{(k)} = [\hat{x}_i, \hat{y}_i, \hat{w}_i, \hat{h}_i] \) represents a bounding box and \( \hat{c}_i^{(k)} \in \{1, \ldots, C\} \) is a class label, given thermal images \( I^{thm, (k)} \in \mathbb{R}^{W \times H} \) and RGB images \( I^{rgb,(k)} \in \mathbb{R}^{W \times H \times 3} \). Distillation training aims to train a smaller student model parameterized by \( \theta_S \) that could closely mimic \( L \) level transformed feature maps \(F_S^{(L)} \) with those \(F_T^{(L)} \) of the teacher model, and to approximate the probabilistic outputs of a larger teacher model, leveraging soft class probabilities and a set of bounding boxes  incorporating ground-truth annotations.
\subsubsection{Knowledge Distillation Loss}

The distillation loss encourages the student to mimic the teacher’s probabilistic outputs \cite{kd_od}, comprising class probability and bounding box components.

\paragraph{Class Probability Distillation}
For each target \( i \), the teacher provides a softened class probability distribution \cite{kd_od} using a temperature \( \tau > 1 \), which can be expressed as
\begin{align}
\textstyle
p_{T,i}^{(k)}(\hat{c}_{T,i}^{(k)} = j | I^{thm,(k)}, I^{rgb,(k)}; \tau) = \frac{\exp(z_{T,j}^{(k)}(i) / \tau)}{\sum_{j'=1}^C \exp(z_{T,j'}^{(k)}(i) / \tau)},
\label{teacher_soft}
\end{align}
where \( z_{T,j}^{(k)}(i) \) is the teacher’s logit for class \( j \) of \(k\)-th sample. The student’s probabilities similarly can be expressed as
\begin{align}
\textstyle
p_{S,i}^{(k)}(\hat{c}_{S,i}^{(k)} = j | I^{thm,(k)}, I^{rgb,(k)}; \theta_S, \tau) = \frac{\exp(z_{S,j}^{(k)}(i; \theta_S) / \tau)}{\sum_{j'=1}^C \exp(z_{S,j'}^{(k)}(i; \theta_S) / \tau)},
\label{student_soft}
\end{align}
The class distillation loss is the Kullback-Leibler (KL) divergence between the softened class probability distributions predicted by the teacher and student models is expressed as
\begin{align}
\textstyle \mathcal{L}_{\text{class-distill}}^{(k)} \!=\! \tau^2 \sum_{i=1}^{N^{(k)}} \text{KL}(p_{T,i}^{(k)}(\hat{c}_{T,i}^{(k)} | \cdot; \tau) || p_{S,i}^{(k)}(\hat{c}_{S,i}^{(k)} | \cdot; \theta_S, \tau)),
\end{align}
where the KL divergence is formulated as
\begin{align}
\textstyle \text{KL}(p_{T,i}^{(k)} || p_{S,i}^{(k)}) = \sum_{j=1}^C p_{T,i}^{(k)}(\hat{c}_{T,i} = j) \log \left( \frac{p_{T,i}^{(k)}(\hat{c}_{T,i}= j)}{p_{S,i}^{(k)}(\hat{c}_{S,i} = j)} \right).
\end{align}

\paragraph{Bounding Box Distillation}
For \(k\)-th sample, the model predicts \( \hat{B}_{i}^{(k)} =  \{\hat{b}_{i}^{(k)}\}_{i=1}^{N^{(k)}}\)
The bounding box distillation loss is of the FFKD-MM-MTD is expressed as
\begin{align}
\textstyle \mathcal{L}_{\text{box-distill}}^{(k)} = \sum_{i=1}^{N^{(k)}} \text{smooth}_{L1}(\hat{B}_{T,i}^{(k)}, \hat{B}_{S,i}^{(k)}),
\end{align}
where \( \hat{B}_{T,i}^{(k)} \) and \( \hat{B}_{S,i}^{(k)} \) is the teacher and student predicted boxes respectively and
\begin{align}\label{eq_smooth}
\text{smooth}_{L1}(x) =
\begin{cases}
0.5 x^2 & \text{if } |x| < 1, \\
|x| - 0.5 & \text{otherwise}.
\end{cases}
\end{align}
Thus, the KD loss of our problem is formulated as
\begin{align} \label{eq_loss_kd}
\mathcal{L}_{\text{knowledge-distill}}^{(k)} = \lambda_\text{cls}  \mathcal{L}_{\text{class-distill}}^{(k)} + \lambda_\text{reg}  \mathcal{L}_{\text{box-distill}}^{(k)},
\end{align}
where \( \lambda_\text{clss}, \lambda_\text{reg} \in [0, 1] \) balances classification and regression components, and are also design optimization parameters.

\subsubsection{Ground-Truth Loss}

The ground-truth loss measures the student’s error relative to true annotations.
The classification loss is characterized as a cross-entropy loss as
\begin{align}
&\mathcal{L}_{\text{class-CE}}^{(k)} = \textstyle - \sum_{i=1}^{N^{(k)}} \sum_{j=1}^C \mathbf{1}\{\hat{c}_{S,i}^{(k)} = j\} \nonumber\\
&\times \log p_{S,i}(\hat{c}_{S,i}^{(k)} = j | I^{thm,(k)}, I^{rgb,(k)}; \theta_S, \tau=1).
\end{align}
The bounding box loss is characterized as a smooth $L1$ loss for bounding box predictions, which is expressed as
\begin{align}
\textstyle \mathcal{L}_{\text{box-reg}}^{(k)} = \sum_{i=1}^{N^{(k)}} \text{smooth}_{L1}(\hat{b}_{S,i}^{(k)} - \mu_S(i; \theta_S)),
\end{align}
where the smooth function is defined in \eqref{eq_smooth}.
Thus, the total ground-truth loss of the FFKD-MM-MTD is formulated as
\begin{align}
\mathcal{L}_{\text{ground-truth}}^{(k)} = \gamma  \mathcal{L}_{\text{class-CE}}^{(k)} + (1 - \gamma)  \mathcal{L}_{\text{box-reg}}^{(k)},
\end{align}
where \( \gamma \in [0, 1] \) balances classification and regression.

\subsubsection{Feature Distillation Loss}
The objective is to have the student model replicate the intermediate representations produced by the teacher model \cite{feature_distillation} as
\begin{equation}
\mathcal{L}_{\text{feature-distill}} = d( F_S^{(L)}(\theta_S), F_T^{(L)}(\theta_T) )
\end{equation}
where \( d \) denotes a distance function of transform features from \(L\) feature map levels. The optimization problem is to learn student parameters \(\theta_S\) that minimize the distance to the teacher parameters \(\theta_T\). 

The considered FFKD-MM-MTD problem is characterized via optimizing the following problem
\begin{subequations}
\label{eq_problem_form}
    \begin{align}
        \underset{\theta_S, \alpha, \beta, \gamma}{\text{minimize}} \quad 
        & \Big( \alpha  \mathcal{L}_{\text{feature-distill}} +
        \beta  \mathcal{L}_{\text{ground-truth}} + \gamma  \mathcal{L}_{\text{knowledge-distill}}
        \Big) \label{eq_obj_func_distillation_feature_out} \\
        \text{subject to} \quad & p_{S,i}(\hat{c}_{S,i} = j) \geq 0 , \sum_{j=1}^C p_{S,i}(\hat{c}_{S,i} = j) = 1, \forall i \label{eq_constraint_a} \\
        & \mu_S(i; \theta_S) \!\in\! [0, W] \!\!\times\!\! [0, H] \!\!\times\!\! [0, W_{\text{max}}] \!\!\times\!\! [0, H_{\text{max}}],\! \forall i, \label{eq_constraint_b} 
    \end{align}
\end{subequations}
where constraint \eqref{eq_constraint_a} is for class probabilities and constraint \eqref{eq_constraint_b} is for Bounding box validity.

\section{Proposed FFKD-based Training Pipeline for MM-MTD}

\begin{figure*}[h!]
    \centering
    \includegraphics[width=\textwidth, height=0.3\textwidth]{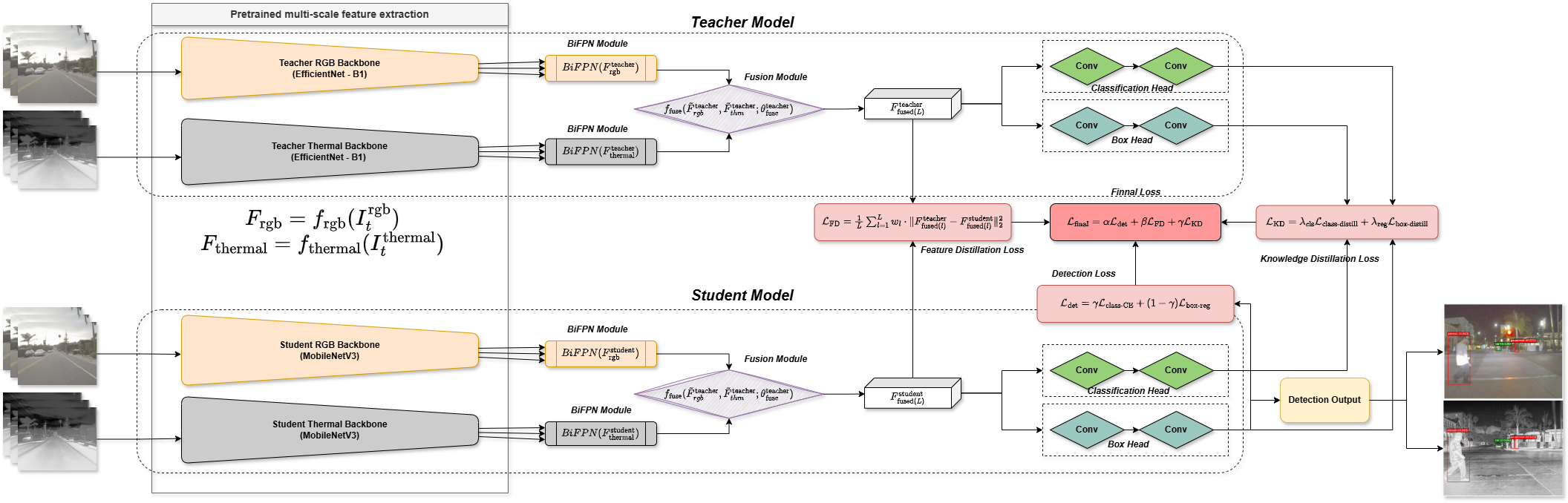}
    \caption{Proposed FFKD-based training pipeline for the considered MM-MTD problem.}
    \label{fig:fig_pipeline}
\end{figure*}

\label{subsec:training_pipeline}

To address the FFKD-MM-MTD optimization problem \eqref{eq_problem_form}, we propose a training algorithm and pipeline that leverages KD to train a lightweight student model for efficient and accurate detection and tracking, as illustrated in Fig \ref{fig:fig_pipeline}. The pipeline processes paired RGB and thermal inputs, employs Bi-Directional Feature Pyramid Networks (BiFPN) \cite{efficientdet} for feature enhancement, ensuring performance on resource-constrained devices. The fused features across modalities by Convolutional Block Attention Module (CBAM) \cite{cbam}, and generate detections through classification and regression heads.

\subsection{Mathematical Description of the Neural Network Architectures for Teacher and Student Models}

To support the Distillation Training-Based pipeline outlined in Section~\ref{subsec:training_pipeline}, we apply a teacher model based on EfficientDet-D1 \cite{efficientdet} and a student model based on MobileNetV3 as the backbone for feature extraction on multi-scale \cite{mobilenetv3}.  

\subsubsection{Teacher Model - EfficientDet-D1}
With the EfficientNet-B1 backbone which uses a method called compound scaling (uniform scalability of the resolution, depth, and width), the teacher model \cite{efficientdet} has parameters \( \theta_{\text{tc}} \), is a high capacity architecture designed for precise detection.

\subsubsection{Student Model - MobileNetV3}
By utilizing hardware-aware network architecture search (NAS) and the NetAdapt algorithm in a complementary way, MobileNetV3 \cite{mobilenetv3} is introduced as a CNNs optimized for mobile CPUs and embedded devices. It introduces novel architecture with parameters \( \theta_{\text{st}} \), including hard swish activation \cite{silu} and squeeze-and-excitation (SE) modules \cite{se} in MBConv blocks, to improve performance.

\subsection{Input Description and Processing}

The pipeline operates on a dataset \( \mathcal{D} = \{(I_t, Y_t)\}_{t=1}^T \). Both models process an input at time \( t \) is \( I_t = \{I_t^{\text{thm}}, I_t^{\text{rgb}}\} \), where \( I_t^{\text{rgb}} \in \mathbb{R}^{H \times W \times 3} \) and \( I_t^{\text{thm}} \in \mathbb{R}^{H \times W \times 1} \) by each backbone modality independently and extract multi-scale features. For each modality \( m \in \{\text{rgb}, \text{thm}\} \):
\begin{align}
F_m = f_{\text{backbone}}(I_t^m; \theta_{\text{backbone}}),
\end{align}
where \( f_{\text{backbone}} \) comprises convolutions blocks, which can be expressed as a composite function as follows
\begin{align}
f_{\text{backbone}} = f_{\text{block}_M} \circ \cdots \circ f_{\text{block}_1},
\end{align}
with each \( f_{\text{block}_m} \) including depthwise separable convolution, point convolution, batch normalization, and swish-based activation \cite{silu}. The output is \( F_m = \{F_m^{l}\}_{l=1}^L \).

These multi-level features are processed through BiFPN modules \cite{efficientdet} with learnable attention weights, which refine and aggregate features across levels iteratively, enhancing cross-scale interactions to improve detection across varying target sizes. The forward propagation of BiFPN is expressed as
\begin{align}
\tilde{F}_m = \text{BiFPN}(F_m).
\end{align}

\subsection{Feature Engineering via Fusion Module}
\label{cbam_fusion}

The refined features are fused across modalities along channel and spatial dimension using a fusion function called CBAM \cite{cbam}, which produces fused feature maps and is parameterized by \( \theta_{\text{fuse}} \). Recall that \(\tilde{F}_{m} \in \{\tilde{F}_{\text{rgb}},\tilde{F}_{\text{thm}} \}\), the data fusion is modeled as
\begin{align}
F_{\text{fused}} = f_{\text{fuse}}(\tilde{F}_{m}; \theta_{\text{fuse}}) \label{teacher_fusion},
\end{align}

\subsection{Final Detection MTD Output}

The fused output features \(F_{\text{fused}}\) are processed by classification and regression heads to produce object bounding boxes and corresponding classes and scores, which are expressed as
\begin{align}
z_{\text{cls}} = f_{\text{cls}}(F_{\text{fused}}; \theta_{\text{cls}}), \quad z_{\text{reg}} = f_{\text{reg}}(F_{\text{fused}}; \theta_{\text{reg}}),
\end{align}
where \( f_{\text{cls}} \),\( f_{\text{reg}} \) are convolutional layers producing classification logits \(z_{\text{cls}} \in \mathbb{R}^{N \times C}\) (for \(N \) anchors and \( C\) classes) and bounding box offsets \(z_{\text{reg}}\). The detection output in \eqref{eq_output_full} can be reformulated as
\begin{align}
D = f_{\text{decode}}(z_{\text{cls}}, z_{\text{reg}}) = \{(\hat{b}_{t,i}, \hat{y}_{t,i})\}.
\end{align}
The comprehensive model of the forward propagation of the FFKD-MM-MTD is mathematically expressed as
\begin{align}
D = &\left( f_{\text{decode}} \circ \left( f_{\text{cls}} \parallel f_{\text{reg}} \right) \right) \circ \notag f_{\text{fuse}} \nonumber\\
&\circ \left( \text{BiFPN} \circ f_{\text{backbone}} \right)^{\text{\text{rgb}},\text{\text{thm}}} \notag (I_t; \theta)
\end{align}

\subsection{The Proposed FFKD-MM-MTD Loss Function Design}
\label{loss_function_design}

To address the objective function \eqref{eq_obj_func_distillation_feature_out}, our training loss combines the actual detection loss of the student model and the transfer loss components which includes the FD loss and the KD loss as
\begin{align}\label{eq_loss_full}
\mathcal{L}_{\text{final}} = \alpha \mathcal{L}_{\text{det}} + \beta \mathcal{L}_{\text{FD}} + \gamma \mathcal{L}_{\text{KD}},
\end{align}
where hyperparameters \( \alpha, \beta, \gamma \) balance the contribution of each loss terms and \(\mathcal{L}_{\text{FD}}\) is applied for the fusion feature maps mentioned in Eq. \eqref{teacher_fusion}.

The training algorithm minimizes \( \mathcal{L}_{\text{final}} \) to optimize \( \theta_{\text{st}} \) of the student model, aligning with the distillation optimization problem in Section~\ref{distillation_realtime_mtd}. Initialized with pre-trained weights, the student parameters are updated over epochs by iterating through \( \mathcal{D} \), computing teacher and student outputs, calculating \( \mathcal{L}_{\text{final}} \), and applying backpropagation update as
\begin{align}
\theta_{\text{st}} \leftarrow \theta_{\text{st}} - \eta \nabla_{\theta_{\text{st}}} \mathcal{L}_{\text{final}},
\end{align}
using an optimizer (e.g., Adam) with learning rate \( \eta \).

\subsection{Inference of the Distillation-Trained Model on New Observations}

Without relying on the teacher, the trained student model takes a new pair of input images and predicts bounding boxes and class labels directly during inference phase. The model outputs the final detections for both RGB and thermal images by applying confidence thresholding and non-maximum suppression techniques.

\section{Results and Discussions}

\subsection{Dataset}
The Teledyne FLIR Thermal Dataset \cite{dataset_flir} provides 26,442 fully annotated thermal and visible spectrum frames to advance multi-target detection and classification for Advanced Driver Assistance Systems (ADAS) and autonomous vehicles. We use the FLIR Aligned Dataset \cite{dataset_flir_aligned} is a refined version of \cite{dataset_flir} that contains 5,142 paired of RGB and thermal images and includes bounding-box annotations for three classes, namely person, car, and bike. The dataset is divided into 4,129 training pairs (for training and validation set) and 1,013 testing pairs.

\subsection{Tranining Setup}
\begin{table}[htbp]
    \centering
    \small
    \caption{Parameter statistic of fusion-based multi-modal models (in millions).}
    \begin{tabular}{lccccc}
        \toprule
        \textbf{Type} & \textbf{Backbone} & \textbf{BiFPN} & \textbf{Head} & \textbf{Total} & \textbf{Trainable} \\
        \midrule
        Teacher & 12.2M & 0.78M & 0.064M & 13.2M & 1M \\
        Student & 3.5M & 0.95M & 0.064M & 4.8M & 1.24M \\
        \bottomrule
    \end{tabular}
    \vspace{2pt}
    \label{tab:param_statistic}
\end{table}
To implement our proposed pipeline, we selected EfficientDet-D1 as the teacher model due to its robust detection performance and balanced accuracy–efficiency trade-off, and MobileNetV3 Large 0.75, a lightweight and computationally efficient architecture, as the student model. Table \ref{tab:param_statistic} presents the statistics of both models. To ensure stable and optimal performance during the training, validation, and testing phases of the proposed method, we conducted a series of experiments to determine appropriate hyperparameters, as shown in our source code.

\begin{figure}[!h]
    \centering
    \includegraphics[width=\linewidth]{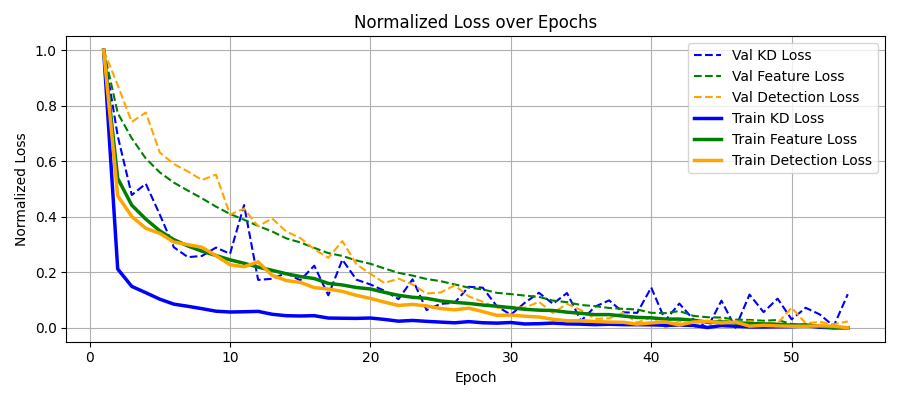}
    \caption{Training and validation loss of each loss component in \eqref{eq_loss_full}.}
    \label{fig:traning_validation_loss}
\end{figure}

\subsection{Inference and Demonstrations}

Figure \ref{fig:traning_validation_loss} shows the convergence of the training process after 50 epochs. To evaluate the effectiveness of our proposed method, we conducted experiments on the test set using mAP at various IoU thresholds. The performance of both teacher and student models under different modality settings is summarized in Tables \ref{tab:map_result} and \ref{tab:inference_time_result}.
\begin{table}[htbp]
    \centering
    \scriptsize
    \caption{Performance on test set of model with mAP where T- prefix is Teacher type model and S- prefix is Student type model}
    \begin{tabular}{l!{\vrule width 0.5pt}ccc}
        \toprule
        \textbf{Model} & \textbf{mAP@0.5:0.95} & \textbf{mAP@0.5} & \textbf{mAP@0.75} \\
        \midrule
        T-RGB Only & 24.4 & 56.2 & 17.2 \\
        T-Thermal Only & 30.7 & 65.2 & 23.7 \\
        T-Fusion & \textbf{33.0} & \textbf{69.1} & \textbf{26.2} \\
        \specialrule{0.5pt}{0.5pt}{0.5pt}
        S-RGB Only & 22.4 & 53.0 & 15.7 \\
        S-Thermal Only & 29.0 & 61.4 & 23.2 \\
        S-Fusion & 27.9 & 58.7 & 22.4\\
        S-Distillation & \textbf{31.5} & \textbf{64.8} & \textbf{24.7} \\
        \bottomrule
    \end{tabular}
    \vspace{2pt}
    \label{tab:map_result}
\end{table}

The table result shows the impact of cross-modal fusion and the effectiveness of knowledge transfer in enhancing significantly smaller student model's performance compared to the single one. Moreover, the S-Distillation model achieves a remarkably high performance, with an index of mAP@0.5:0.95 at 31.5\%, which is comparable to the teacher fusion model (T-Fusion) at 33.0\%.

\begin{table}[htbp]
    \centering
    \scriptsize
    \caption{Performance on test set of model with inference time}
    \begin{tabular}{lccc}
        \toprule
        \textbf{Model} & \textbf{Batch Size} & \textbf{Inference Speed (s)} \\
        \midrule
        T-Fusion & 32 & 0.041 \\
        S-Distillation & 32 & 0.023 \\
        \bottomrule
    \end{tabular}
    \vspace{2pt}
    \label{tab:inference_time_result}
\end{table}

\begin{figure}
\centering
    \subfloat[]{
        \includegraphics[width=\linewidth]{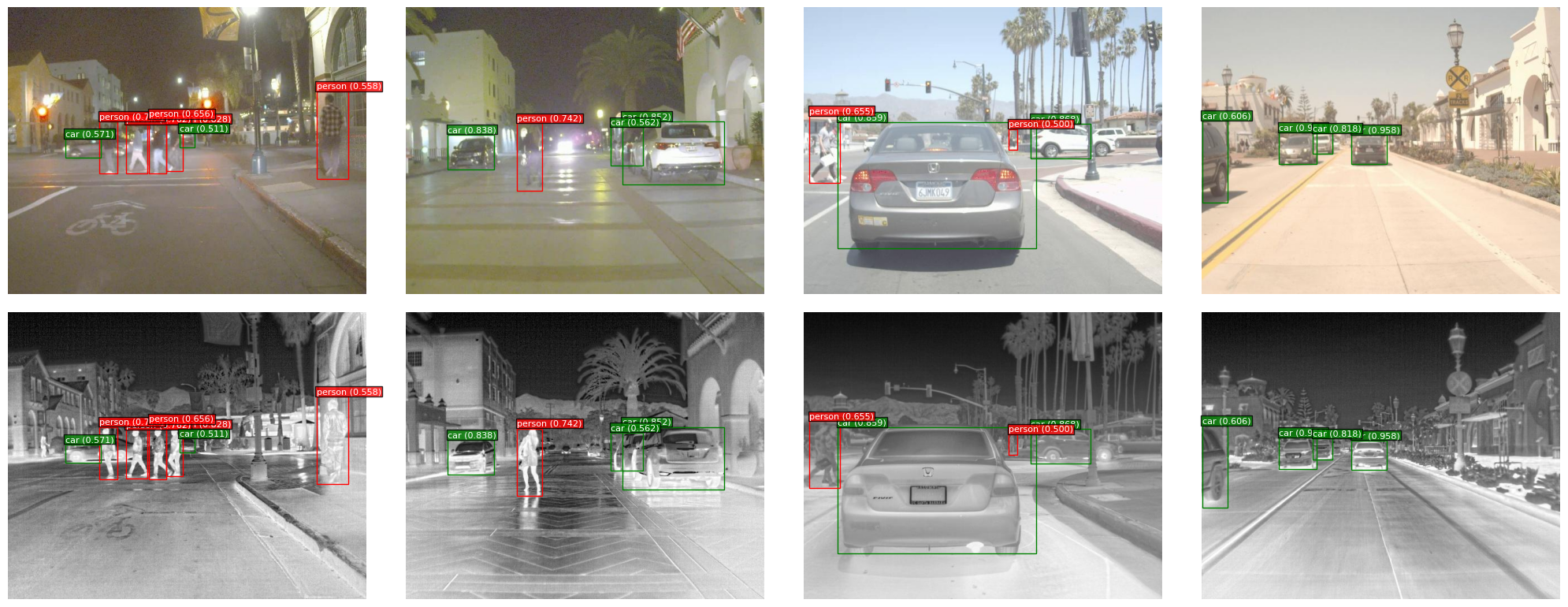}
        \label{fig_map}
    }\\
    \subfloat[]{
        \includegraphics[width=\linewidth]{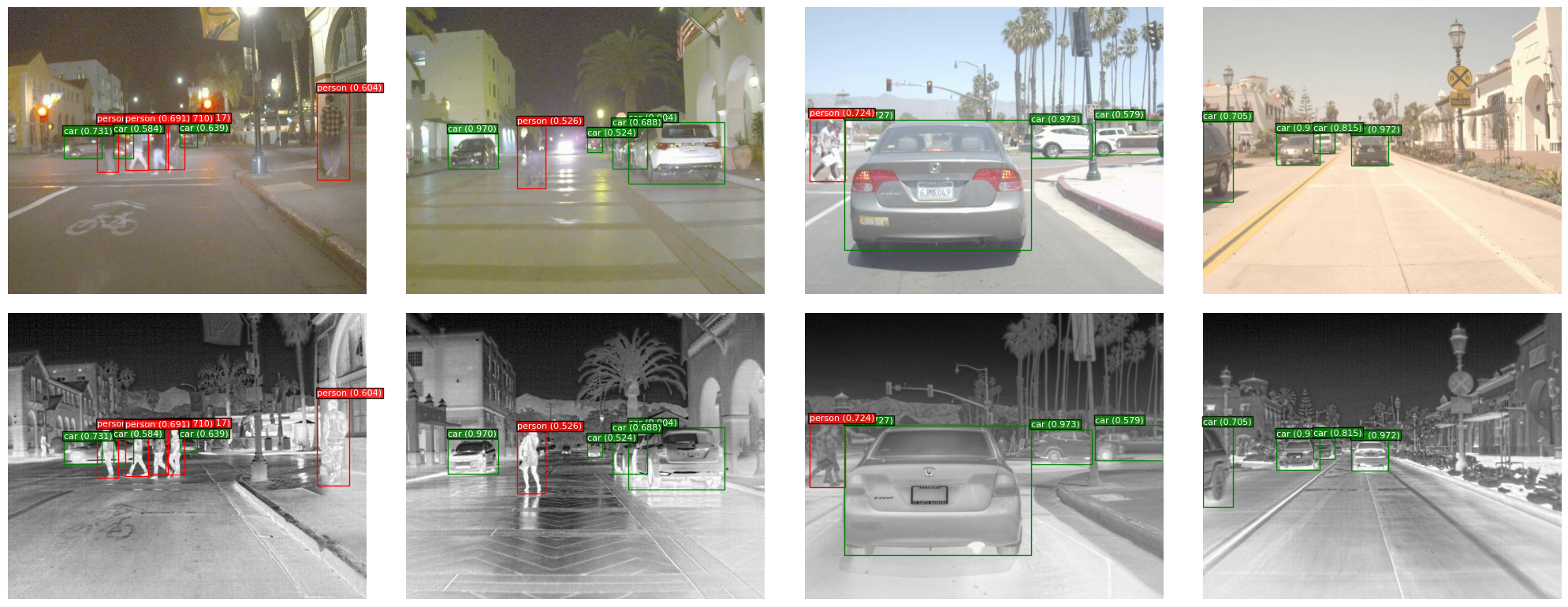}
        \label{fig:teacher_predict_images}
    }
\caption{Predicted Images from (a) teacher model and (b) student model. Please refer to the code repository for further demonstrations.}
\label{fig:predict_images}
\end{figure}

The S-Distillation model's inference speed is computational efficiency achieving a significantly faster of 0.023 seconds per image and is approximately 50\% of the inference speed of the T-Fusion model. A demonstration of the qualitative predicted images from both the fusion-based teacher model and the distilled student model is provided in Fig.~\ref{fig:predict_images}. in which there are no significant differences in detection quality.

\section{Conclusion}
We address the problem of multi-modal multi-target detection and classification (MM-MTD) in autonomous driving using the realistic FLIR dataset. We propose a novel training pipeline that integrates fusion-based multi-modal modeling with knowledge distillation to develop an efficient and compact model for MTD. Our solution, termed as FFKD-MM-MTD pipeline, introduces a principled optimization formulation and a tailored composite loss function for the detection task. Experimental results demonstrate that our student model achieves approximately 95\% of the teacher model’s mAP performance while reducing inference time by approximately 50\%, facilitating deployment on resource-constrained edge devices.

\bibliographystyle{IEEEtran}
\bibliography{References}

\end{document}